\documentclass[twocolumn, pra, showpacs, aps, superscriptaddress]{revtex4-1}

\usepackage{amsmath}
\usepackage{amsfonts}
\usepackage{bbold}

\usepackage{graphicx}
\usepackage{ae,aecompl}
\usepackage{bm}
\usepackage{textcomp}
\usepackage{xcolor}

\usepackage{wasysym}
\usepackage{hyperref}

\usepackage{natbib}

\begin{document}


\date{\today}

\title{Ice Core Science Meets Computer Vision: Challenges and Perspectives}

\author{P. Bohleber}
\email{pascal.bohleber@unive.it}
\affiliation{Department of Environmental Sciences, Informatics and Statistics, Ca’Foscari University of Venice, Italy}
\author{M. Roman}
\affiliation{Department of Environmental Sciences, Informatics and Statistics, Ca’Foscari University of Venice, Italy}
\author{C. Barbante}
\affiliation{Department of Environmental Sciences, Informatics and Statistics, Ca’Foscari University of Venice, Italy}
\affiliation{Institute of Polar Sciences, CNR, Venice, Italy}
\author{S. Vascon}
\affiliation{Department of Environmental Sciences, Informatics and Statistics, Ca’Foscari University of Venice, Italy}
\author{K. Siddiqi}
\affiliation{School of Computer Science \& Centre for Intelligent Machines, McGill University, Montreal, Canada}
\author{M. Pelillo}
\affiliation{Department of Environmental Sciences, Informatics and Statistics, Ca’Foscari University of Venice, Italy}

\begin{abstract}
Polar ice cores play a central role in studies of the earth's climate system through natural archives. A pressing issue is the analysis of the oldest, highly thinned ice core sections, where the identification of paleoclimate signals is particularly challenging. For this, state-of-the-art imaging by laser-ablation inductively-coupled plasma mass spectrometry (LA-ICP-MS) has the potential to be revolutionary due to its combination of micron-scale 2D chemical information with visual features. However, the quantitative study of record preservation in chemical images raises new questions that call for the expertise of the computer vision community. To illustrate this new inter-disciplinary frontier, we describe a selected set of key questions. One critical task is to assess the paleoclimate significance of single line profiles along the main core axis, which we show is a scale-dependent problem for which advanced image analysis methods are critical. Another important issue is the evaluation of post-depositional layer changes, for which the chemical images provide rich information. Accordingly, the time is ripe to begin an intensified exchange among the two scientific communities of computer vision and ice core science. The collaborative building of a new framework for investigating high-resolution chemical images with automated image analysis techniques will also benefit the already wide-spread application of LA-ICP-MS chemical imaging in the geosciences. 

\section*{Keywords:} ice core analysis, chemical imaging, laser ablation inductively-coupled plasma mass spectrometry, image analysis, scale spaces, co-localization, grain boundaries
\end{abstract}
\maketitle 

\section{Introduction}
Ice cores drilled on the polar ice sheets are among the most important climate archives, delivering valuable insights into the complexity of our climate system \citep{fischer2021ice}. With the deposition of snow, additional properties of the atmospheric composition (aerosols, stable water isotopes, etc.) are archived. When snow is transformed to ice, air bubbles are enclosed, providing actual snapshots of the atmosphere of the past and constituting a stand-alone feature of this natural archive. As a result, there exists a chemical and structural stratification of polar ice, which make up the paleoclimatic record. The layering encoding this record is subject to changes with time and depth, as it is not independent from its conserving matrix. In fact, the interaction between the paleoclimatic record and its preserving ice matrix extends from km-scale ice flow over millennia, down to the microscopic scale of interactions between impurities and ice crystals \citep{faria2010polar}.

As a polycrystal, glacier ice generally consists of a large number of individual crystals (grains), as well as associated microscopic features such as grain boundaries and triple junctions at their intersection. The microscopic localization of impurities with respect to the configuration of grains and their boundaries can crucially affect not only macroscopic deformational properties, but also the paleoclimate records: For instance, diffusion and segregation of soluble impurities into grain boundaries may alter or even destroy the sequence of paleoclimate signals  \citep{rempel2001possible,barnes2004distribution,ng2020pervasive}. Therefore, a quantitative study of the ice stratigraphy at the microscale, in particular the 2D distribution of chemical impurities, is crucial to understand how climate signals have been conserved or transformed, within the flowing ice over tens to hundreds of millennia. This is of particular relevance for the  retrieval of novel climate records from the deepest, oldest and thinnest layers, the primary target of the upcoming ambitious “Oldest ice” ice core drillings in Antarctica, which is a grand challenge in the ice core sciences \citep{brook2006future,fischer2013find}. 

Ice core stratigraphy can be assessed through a variety of methods. The most common approach is direct visual inspection, combined with microstructural analysis of selected ice sections via optical microscopy. Such datasets have revealed new insights into the complexity of the underlying physical processes, challenging established glaciological concepts \citep{kipfstuhl2009evidence, faria2014microstructure}.
Initial contact points with automated image analysis have now been developed. Early studies used coaxial reflected light and the birefringed properties of ice in visual images to automatically detect grain boundaries via image segmentation \citep{arnaud1998imaging, gay1999automatic}. For datasets from directed bright-field illumination, a machine learning approach was developed to obtain automatic, reliable classification of ice crystal features \citep{binder2013extraction}. At the microscale, a digital form of optical microscopy was combined with automated image analysis to quantify the collective grain as well as bubble structure  \citep{kipfstuhl2009evidence, ueltzhoffer2010distribution, bendel2013high}. \citet{morcillo2020unravelling} recently employed digital image analysis for a macroscopic counterpart of dark-field microscopy to investigate the optical stratigraphy of an Antarctic ice core.

To fully assess the preservation of the climatic signals represented by the vertical distribution of impurity concentration in ice cores, jointly evaluating the chemical information with visual analysis is key. However, all techniques based on the chemical analysis of ice core meltwater lack the ability to directly assess lateral spatial relations among impurities within the ice, and face fundamental limitations in depth resolution \citep[e.g.][]{breton2012quantifying}. To analyze samples in the frozen ice state, scanning electron microscopy/energy dispersive X-ray spectrometry \citep[e.g.][]{barnes2003sem, baker2003sem, iizuka2009constituent} as well as micro-Raman spectroscopy \citep[e.g.][]{eichler2019impurity} have been successfully used to determine the location of impurities within the ice matrix and to analyze the chemical composition of individual particles, typically based on spot-like measurements. To evaluate the chemical micro-stratigraphy and its implications for the interpretation of coarser scale meltwater signals calls for continuous 2D information. However, a standard technique in 2D chemical imaging has not yet been established for ice cores. 

A highly promising candidate in this regard is laser ablation inductively-coupled plasma mass spectrometry (LA-ICP-MS). LA-ICP-MS uses a laser to ablate a few tenths of micro-liters of ice from the surface, which are subsequently analyzed for various elemental impurity species with a mass spectrometer \citep{muller2011direct, sneed2015new, della2017calibrated, spaulding2017new}. Very recently, the application of LA-ICP-MS for 2D chemical imaging in ice cores has been refined and greatly improved \citep{bohleber2020imaging}. Here high-resolution artifact-free 2D images from state-of-the-art imaging methods \citep[e.g.][]{wang2013fast, van2019insights} provide an unprecedented density of information; several million laser shots (correlating with chemical measurements) are fired over just a few square mm. A camera co-aligned with the laser then captures visual images of the ice sample surface. By this means, LA-ICP-MS combines chemical and visual information of the ice core microstratigraphy.

The introduction of a new technology establishes the need for better understanding the potential and limitations of these novel high-resolution 2D chemical images of ice cores. As we see new frontiers of multidisciplinary research emerging in this framework, the employment of computer vision methods might be a game-changer for fully exploiting the complexity of the chemical images. In the same fashion as computer vision has lead to advances in medical image analysis, neuroscience, bio-medicine, robotics, object recognition and a host of other areas, the time is ripe for it to revolutionize ice core science. To accelerate this, in this article we begin an inter-disciplinary dialogue. We describe a selected set of key problems in the analysis of chemical images of ice cores as examples of future inter-disciplinary research questions, that can only be successfully tackled in close collaboration with the computer vision community.

\section{Making chemical images of ice cores}
\begin{figure*}[ht]
\centering
\includegraphics[width=.65\textwidth]{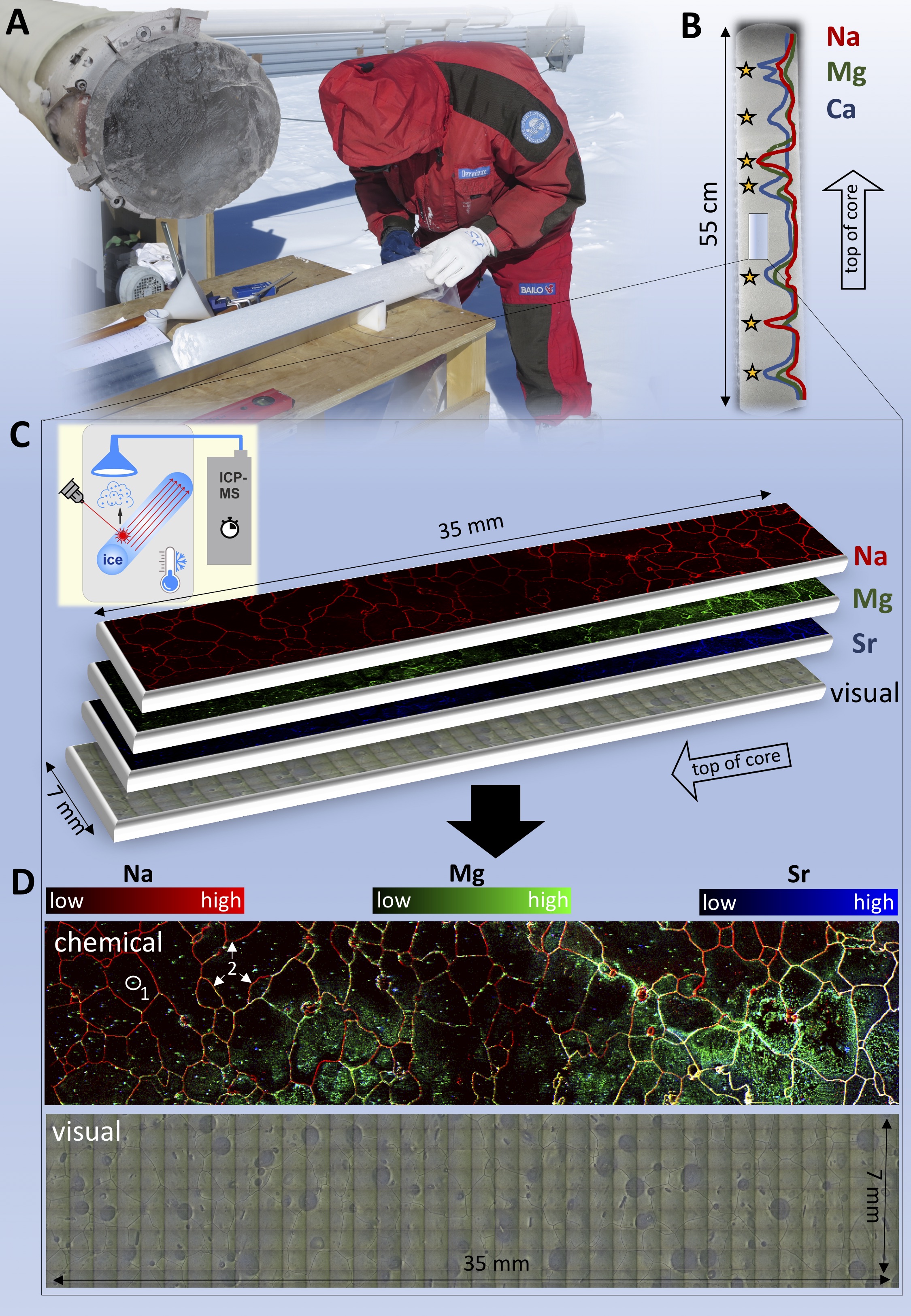}
\caption{LA-ICP-MS imaging at micron-scale resolution combines new chemical dimensions with visual ice core analysis. Shown in A is an ice core drill and a freshly drilled ice core piece. Typical cm-resolution signals of chemical impurities obtained from meltwater analysis are illustrated in B (photo credit: Sarah Wauthy). Stars indicate hypothetical stratigraphic (e.g. annual) layers. Row C shows the  chemical imaging with LA-ICP-MS, with examples for the chemical image for Na, Mg and Sr intensities, using separate color channels and the visual images obtained from a mosaic of camera images (in D). The comparison shows individual bright spots (1) and that the network of lines most dominant in the red Na image (2) correspond to the grain boundaries which are seen as dark lines in the visual image.}
\label{fig:1}
\end{figure*}

Impurity images are acquired as a pattern of lines, without overlap in the direction perpendicular to that of the scan, and without any further spatial interpolation. Each pixel in an ice core chemical image has a size of 35 µm x 35 µm (Figure \ref{fig:1}). For each chemical element a numerical matrix contains rows and columns, according to the physical size of the image: an image of 7 mm x 35 mm in size has 200 rows and 1000 columns. The numerical entries in this matrix refer to either the recorded intensity (e.g. counts per second) or, if a calibration is applied, the concentration (e.g. in parts per million). Due to the careful synchronization, the individual pixels of the different chemical channels can be considered to be almost perfectly spatially aligned. In contrast, the mosaic of visual images obtained from the laser camera is not a-priori aligned with the chemical images. A sample dataset (including the datasets from Figure \ref{fig:1}) has been made publicly available (see data availability below). Details on the imaging technique are described elsewhere \citep{bohleber2020imaging}. 

The visual images are generally characterized by air bubbles (dark blobs), grain boundaries (dark lines) and occasional sub-grain boundaries (thin dark lines). In the chemical images, the individual impurity channels generally differ, e.g. through a variable degree of impurity localization at grain boundaries, depending on the depth sections and respective climatic period. The LA-ICP-MS image shown in Figure \ref{fig:1} has been obtained from the EPICA Dome C deep Antarctic ice core, on a sample from a glacial period of about 27.8 thousand years ago. Additional images are presented elsewhere \citep{bohleber2020two}. From a conceptual point of view, Figure \ref{fig:1} is a representative example of the following general image characteristics:
i) Small bright spots of a few ten microns occurring within grains as single spots or in small clusters. “Clouds” of such spots are also observed.
ii) High intensities reflected in co-localization with the grain boundary network, especially for the Na images. In a typical 7 mm x 35 mm image, tens of grains to a few hundred grains can be distinguished visually.
iii) Overall intensity gradients on the mm-scale, showing sections with darker versus brighter intensities.

\section{Key questions and a role for computer vision}
The future technological evolution of LA-ICP-MS ice core analysis promises to increase image size by at least one order of magnitude or more, enabled by large cryocells \citep{sneed2015new} and faster scan speeds \citep{vsala2020analytical}. However, in order to keep pace with this expected technological breakthrough an equal advance in data reduction and interpretation is called for. Below we discuss three crucial questions related to signal identification and record preservation, where the application of computer vision methods could lead to breakthroughs.

\subsection*{Question 1: Can scale-space theory assist in extracting paleoclimatic signals?}

Previous investigations focused on establishing a meaningful interpretation of the novel LA-ICP-MS signals through a validation against existing meltwater analysis. It was shown that the coarse-scale variability seen in 1D single line profiles of LA-ICP-MS, measured along the main core axis, is consistent with the full resolution signals from meltwater analysis \citep{della2017calibrated, spaulding2017new}. The coarse-scale variability had to be inferred by applying substantial (e.g. Gaussian) smoothing to 1D single line profiles. In this fashion annual layers could be identified (i.e. the chemical contrast between summer and winter snow layers) which were thinned beyond the cm-resolution capabilities of meltwater analysis \citep{bohleber2018temperature}. However, smoothing with a Gaussian simply discards much of the high-resolution signal, and crucial information is potentially lost.

Initial evidence has already suggested a relationship between the fine-scale signal components (e.g. peaks in 1D line profiles) and the grain boundaries \citep{della2014location, kerch2015laser, beers2020triple}. The new 2D imaging technique has fully revealed the influence of ice crystal features and has demonstrated that, if an impurity species is mostly localized at the grain boundaries (such as Na), it is in turn the grain boundary network that determines the fine-scale signal components in single line profiles \citep{bohleber2020imaging}. Accordingly, there should exist a hypothetical ``sweet spot'' in measurement scale, at which the stratigraphic record (e.g. annual layers) is captured at the highest possible resolution, while the imprint of crystal-related features such as grain boundaries is still avoided. 

However, the physical scales of stratigraphic layering and ice crystal features are neither constant in depth nor universal among all cores: the stratigraphic layering will become increasingly compressed due to deformation by ice flow. The size of ice grains also varies between glacial and interglacial periods, and can become increasingly large in deep ice due to several processes. Eventually sub-mm (annual) layer thickness can be expected in ice, showing grain diameters exceeding several cm. This means that the localization of impurities at grain boundaries can ultimately erase the original layer sequence. In this case the crystal size becomes an upper limit to the resolution at which the related paleoclimate signals may be interpreted.  Accordingly, to separate the imprint of ice crystal features from the stratigraphic layering may not be achievable in all instances, especially once the stratigraphic signal gets corrupted by layer folding, impurity segregation to grain boundaries and migration along the ice vein network. 

\begin{figure*}[ht]
\includegraphics[width=.95\textwidth]{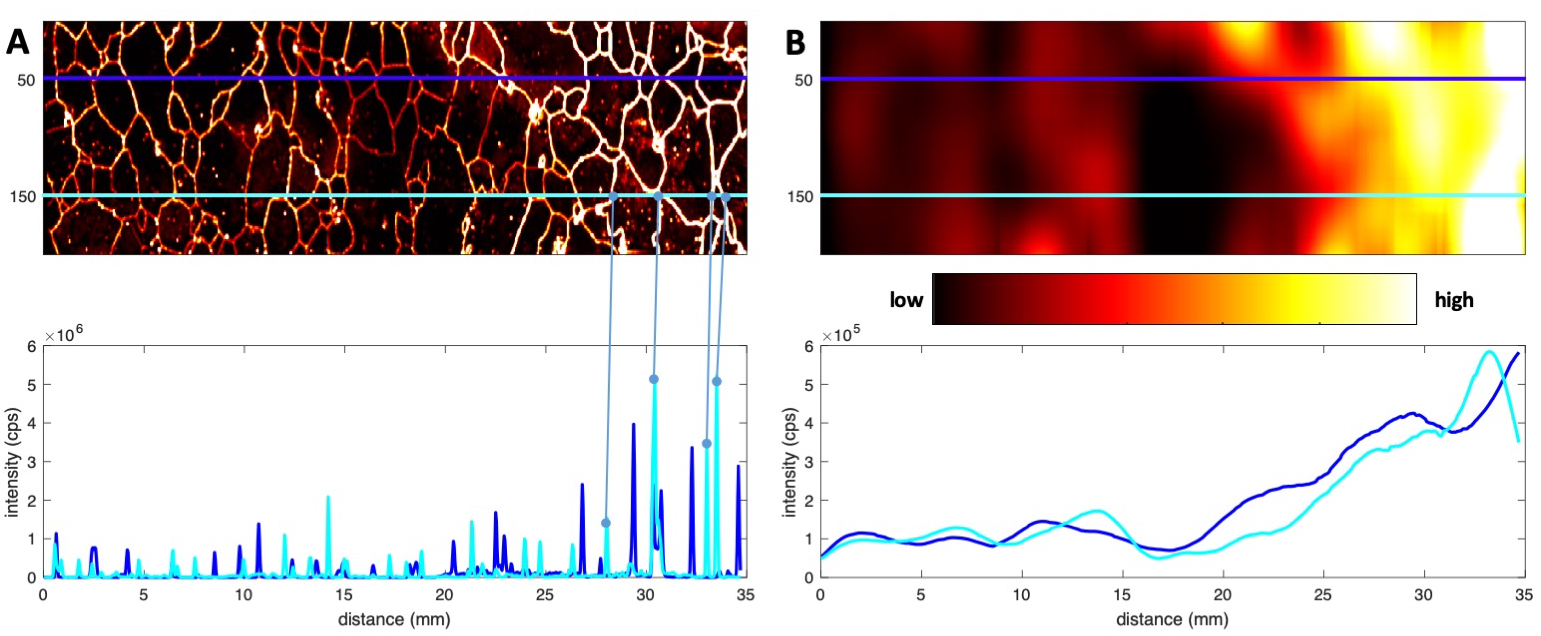}
\caption{Image A shows the original Na image (7 mm x 35 mm). The example of two individual line profiles (the dark and light blue lines, two of the 200 horizontal lines that make up this image) reveals how the significance of single peaks (the fine scale signal components) is tied closely to the grain boundaries in the ice crystal network (illustrated by four sample connecting lines). The imprint of the grain boundaries on the LA-ICP-MS signals weakens if the resolution is decreased and the scale is increased, respectively. This is illustrated by image B, obtained by applying 2D Gaussian smoothing to image A (30 pixel standard deviation). In image B, the two lines spaced 100 pixels (350 µm) apart show a high degree of similarity, the coarse scale components now representing mainly the overall intensity gradient of the image.}
\label{fig:2}
\end{figure*}

As a starting point to tackle this question, undisturbed ice core sections typically feature a nearly level layer sequence \citep{svensson2005visual}. In this case, signals related to the stratigraphic layering should not be a primary function of the lateral position of the line profile on the ice core surface \citep{sneed2015new, della2017calibrated}. However, due to the presence of the grain boundary imprint, this does not hold at the high resolution scale, as shown in Figure \ref{fig:2}. On the other hand, for the example of the Na image of Figure \ref{fig:1}, the grain boundary imprint becomes increasingly blurred until, at a scale of around 1 mm (e.g. the standard deviation of the Gaussian kernel), only the overall intensity gradients of the image remain present. This suggests that the overall intensity gradient is connected to the stratigraphic layering, warranting further investigation. Small-scale folding would corrupt the above criterion, with regard to the lateral signal coherence. An answer cannot come from 1D line profiles alone, but rather requires an analysis of the 2D images.

Evidently, this question has a deep connection to the scale of the image features. Hence, much may be gained from treating this question with scale-space techniques, a well-established field in computer vision \citep{witkin1983scale,Koe84,Lin94,florack1994linear,florack1996gaussian,ter2003front,duits2004axioms, Koenderink2021} for which modern approaches leveraging deep learning are now emerging \citep{bekkers2018roto,romero2020attentive}. The expected output from scale-dependent image characterization should allow one  to optimize the experimental design to reliably detect the stratigraphic layer signal at the highest possible resolution. This can mean determining the physical resolution of analysis, such as the laser spot size, and determining the number of and lateral spacing between the parallel lines needed - ideally including a criterion for the detection of small-scale folding. 

\subsection*{Question 2: What do image features reveal about the ice core record?}

Detecting, counting and classifying grain boundaries, triple junctions and spot-type image features along the chemical image dimensions can offer new and improved way of extracting chemical information. For instance, marine and terrestrial sources of insoluble particles have been distinguished based on a classification according to the multi-elemental composition \citep{oyabu2020compositions}. For the chemical images such a classification is a complex task that needs to be performed automatically, or it will be intractable with growing image size. Thus, this is another area where expertise from computer vision will be instrumental in overcoming the present limitations.

The chemical images also offer new options for the investigation of post-depositional chemical reactions. As a concrete example in this regard, meltwater analysis of a deep Antarctic ice core revealed that sharp sulfate spikes showed an ``anomalous'' chemical composition, which was explained by the post-depositional formation of soluble particles of magnesium sulfate salts in grain boundaries \citep{traversi2009sulfate}. The latter process would imply the co-localization of Mg and Ca in grain boundaries. Chemical images can provide new insights about whether the \textit{co-occurrence} of impurities in meltwater analysis is in fact a result of their \textit{co-localization} at grain boundaries. Since the chemical images will, for the foreseeable future, only cover a small representative section of a typical sample melted for analysis (e.g. approximately 1 m long, cross-section typically 3 cm × 3 cm \citep{rothlisberger2000technique}) a meaningful way of extrapolation and upscaling is needed when attempting a direct comparison to meltwater techniques. In this context, macroscopic variables derived from the images can be compared more easily to those of other datasets. Examples of such variables are grain size and shape as well as the spatial density and chemical signature of insoluble particles (related to spot-type features). This can be compared, e.g. to visual datasets, volume particle concentration derived from macroscopic meltwater analysis and to the glacio-chemical characterization of other techniques. 

Using image features \citep{zhou2015cnnlocalization} in concert with tools including co-localization analysis, image segmentation \citep{activecontours, malladi1995shape,chan2001active, FCN15, UNET} clustering \citep{NormalizedCut,spectralclustering,pavan2007dominant} and object detection \citep{FRCNN} can provide a foundation for performing such analysis in a quantitative manner. For instance, to assess the layer integrity, an algorithm must determine i) the degree of localization of an impurity species at the grain boundaries and weigh ii) the average size of the grains against iii) the thickness of the stratigraphic layering of interest. The latter data (expected nominal thickness of annual layers etc.) can typically come from existing datasets from glaciological and ice core analyses. 

\subsection*{Question 3: Can computer vision help to test existing theories of impurity diffusion?}

Although this third problem may not be solvable based on the images thus far available, it is of central relevance to ongoing and future efforts to retrieve paleoclimate information from deep ice. Diffusion of impurities along ice veins (triple junctions) due to concentration gradients of soluble species or chemical competition leads to the broadening of peak-like signals \citep{barnes2004distribution}. There is also the case of ``anomalous diffusion'' which can lead to peak displacement along the main core axis \citep{rempel2001possible}. In a recent theoretical advance it has been shown that diffusive impurity transport along ice veins will lead to an effect much like low-pass filtering on 1-D signals \citep{ng2020pervasive}. 

Whether the diffusive becomes strong enough to affect the paleoclimate signals evidently depend on two factors: i) the relative degree of localization at triple junction (i.e. most of the impurities need to be located there) and ii) the connectedness of the ice vein network (vs. blockages or void areas). In principle, both of these factors can be constrained through the chemical images. The degree of connectedness of the ice vein network could be evaluated using a 3D approach. For this purpose, 2D images are stacked together, either in consecutive ablation or after removing a surface layer manually by scraping/polishing the surface \citep{van2017imaging}. The latter choice appears more realistic considering that the ice layer ablated by a single image has a thickness in the micron-range. This means that building a 3D model of the ice vein network may require some interpolation or statistical treatment of all detected triple junctions. 

\section{Outlook}
Expertise in computer vision, pattern recognition and machine learning will be critical to exploit the full potential of the chemical imaging for addressing the key challenges in deep Antarctic ice. In the short term, the examples presented here already highlight valuable starting points. These concern automatically capturing statistical information and carrying out chemical classification, and predicting the scale at which the stratigraphic layering may be preserved. As the chemical images continue to grow in size, so will the urgency to address these issues. 
Regarding a vision for a prolonged long term partnership between ice core and computer science, it can be envisaged that, once initial algorithms have been developed, a next step could involve further automation, e.g. to train a machine learning algorithm to recognize instances of stratigraphic layer disturbances in the chemical images. This would result in a similar machine-based assistance as in state-of-the-art medical diagnostics. In this way, computer vision has the potential to lead to a permanent transformation in the manner that chemical stratigraphy in polar ice cores is investigated.

\section*{Conflict of Interest Statement}
The authors declare that the research was conducted in the absence of any commercial or financial relationships that could be construed as a potential conflict of interest.

\section*{Author Contributions}
All authors contributed to the discussion and the writing of the manuscript. PB, MR and CB established the ice core chemical imaging. MP conceived the original idea behind this manuscript. 

\section*{Funding}
PB gratefully acknowledges funding from the European Union's Horizon 2020 research and innovation programme under the Marie Skłodowska-Curie grant agreement No. 790280. KS is grateful to the Natural Sciences and Engineering Research Council of Canada (NSERC) for research funding. This publication was generated in the frame of Beyond EPICA. The project has received funding from the European Union's Horizon 2020 research and innovation programme under grant agreement No. 815384 (Oldest Ice Core). It is supported by national partners and funding agencies in Belgium, Denmark, France, Germany, Italy, Norway, Sweden, Switzerland, The Netherlands and the United Kingdom. Logistic support is mainly provided by PNRA and IPEV through the Concordia Station system. The opinions expressed and arguments employed herein do not necessarily reflect the official views of the European Union funding agency or other national funding bodies. This is Beyond EPICA publication number XX.

\section*{Acknowledgments}
The authors thank Ciprian Stremtan and Alessandro Bonetto for their continued technical support.

\section*{Data Availability Statement}
The underlying datasets presented in this study will be made available via a public data repository (Pangaea) after completion of the peer-review process.


\bibliographystyle{apalike}
\bibliography{Bohleber_etal_Arxiv}

\begin{thebibliography}{}

\bibitem[Arnaud et~al., 1998]{arnaud1998imaging}
Arnaud, L., Gay, M., Barnola, J.-M., and Duval, P. (1998).
\newblock Imaging of firn and bubbly ice in coaxial reflected light: a new
  technique for the characterization of these porous media.
\newblock {\em Journal of Glaciology}, 44(147):326--332.

\bibitem[Baker and Cullen, 2003]{baker2003sem}
Baker, I. and Cullen, D. (2003).
\newblock Sem/eds observations of impurities in polar ice: artifacts or not?
\newblock {\em Journal of Glaciology}, 49(165):184--190.

\bibitem[Barnes and Wolff, 2004]{barnes2004distribution}
Barnes, P.~R. and Wolff, E.~W. (2004).
\newblock Distribution of soluble impurities in cold glacial ice.
\newblock {\em Journal of Glaciology}, 50(170):311--324.

\bibitem[Barnes et~al., 2003]{barnes2003sem}
Barnes, P.~R., Wolff, E.~W., Mallard, D.~C., and Mader, H.~M. (2003).
\newblock Sem studies of the morphology and chemistry of polar ice.
\newblock {\em Microscopy research and technique}, 62(1):62--69.

\bibitem[Beers et~al., 2020]{beers2020triple}
Beers, T.~M., Sneed, S.~B., Mayewski, P.~A., Kurbatov, A.~V., and Handley,
  M.~J. (2020).
\newblock Triple junction and grain boundary influences on climate signals in
  polar ice.
\newblock {\em arXiv preprint arXiv:2005.14268}.

\bibitem[Bekkers et~al., 2018]{bekkers2018roto}
Bekkers, E.~J., Lafarge, M.~W., Veta, M., Eppenhof, K.~A., Pluim, J.~P., and
  Duits, R. (2018).
\newblock Roto-translation covariant convolutional networks for medical image
  analysis.
\newblock In {\em International conference on medical image computing and
  computer-assisted intervention}, pages 440--448. Springer.

\bibitem[Bendel et~al., 2013]{bendel2013high}
Bendel, V., Ueltzh{\"o}ffer, K.~J., Freitag, J., Kipfstuhl, S., Kuhs, W.~F.,
  Garbe, C.~S., and Faria, S.~H. (2013).
\newblock High-resolution variations in size, number and arrangement of air
  bubbles in the epica dml (antarctica) ice core.
\newblock {\em Journal of Glaciology}, 59(217):972--980.

\bibitem[Binder et~al., 2013]{binder2013extraction}
Binder, T., Garbe, C.~S., Wagenbach, D., Freitag, J., and Kipfstuhl, S. (2013).
\newblock Extraction and parametrization of grain boundary networks in glacier
  ice, using a dedicated method of automatic image analysis.
\newblock {\em Journal of microscopy}, 250(2):130--141.

\bibitem[Bohleber et~al., 2018]{bohleber2018temperature}
Bohleber, P., Erhardt, T., Spaulding, N., Hoffmann, H., Fischer, H., and
  Mayewski, P. (2018).
\newblock Temperature and mineral dust variability recorded in two
  low-accumulation alpine ice cores over the last millennium.
\newblock {\em Climate of the Past}, 14(1):21--37.

\bibitem[Bohleber et~al., 2020a]{bohleber2020imaging}
Bohleber, P., Roman, M., {\v{S}}ala, M., and Barbante, C. (2020a).
\newblock Imaging the impurity distribution in glacier ice cores with
  la-icp-ms.
\newblock {\em Journal of Analytical Atomic Spectrometry}, 35(10):2204--2212.

\bibitem[Bohleber et~al., 2020b]{bohleber2020two}
Bohleber, P., Roman, M., {\v{S}}ala, M., Delmonte, B., Stenni, B., and
  Barbante, C. (2020b).
\newblock Two-dimensional impurity imaging in deep antarctic ice cores:
  Snapshots of three climatic periods and implications for high-resolution
  signal interpretation.
\newblock {\em The Cryosphere Discussions}, pages 1--21.

\bibitem[Breton et~al., 2012]{breton2012quantifying}
Breton, D.~J., Koffman, B.~G., Kurbatov, A.~V., Kreutz, K.~J., and Hamilton,
  G.~S. (2012).
\newblock Quantifying signal dispersion in a hybrid ice core melting system.
\newblock {\em Environmental science \& technology}, 46(21):11922--11928.

\bibitem[Brook et~al., 2006]{brook2006future}
Brook, E.~J., Wolff, E., Dahl-Jensen, D., Fischer, H., Steig, E.~J., et~al.
  (2006).
\newblock The future of ice coring: international partnerships in ice core
  sciences (ipics).
\newblock {\em PAGES news}, 14(1):6--10.

\bibitem[Chan and Vese, 2001]{chan2001active}
Chan, T.~F. and Vese, L.~A. (2001).
\newblock Active contours without edges.
\newblock {\em IEEE Transactions on image processing}, 10(2):266--277.

\bibitem[Della~Lunga et~al., 2014]{della2014location}
Della~Lunga, D., M{\"u}ller, W., Rasmussen, S.~O., and Svensson, A. (2014).
\newblock Location of cation impurities in ngrip deep ice revealed by cryo-cell
  uv-laser-ablation icpms.
\newblock {\em Journal of Glaciology}, 60(223):970--988.

\bibitem[Della~Lunga et~al., 2017]{della2017calibrated}
Della~Lunga, D., M{\"u}ller, W., Rasmussen, S.~O., Svensson, A., and
  Vallelonga, P. (2017).
\newblock Calibrated cryo-cell uv-la-icpms elemental concentrations from the
  ngrip ice core reveal abrupt, sub-annual variability in dust across the
  gi-21.2 interstadial period.
\newblock {\em The Cryosphere}, 11(3):1297--1309.

\bibitem[Duits et~al., 2004]{duits2004axioms}
Duits, R., Florack, L., De~Graaf, J., and ter Haar~Romeny, B. (2004).
\newblock On the axioms of scale space theory.
\newblock {\em Journal of Mathematical Imaging and Vision}, 20(3):267--298.

\bibitem[Eichler et~al., 2019]{eichler2019impurity}
Eichler, J., Weikusat, C., Wegner, A., Twarloh, B., Behrens, M., Fischer, H.,
  H{\"o}rhold, M., Jansen, D., Kipfstuhl, S., Ruth, U., et~al. (2019).
\newblock Impurity analysis and microstructure along the climatic transition
  from mis 6 into 5e in the edml ice core using cryo-raman microscopy.
\newblock {\em Frontiers in Earth Science}, 7:20.

\bibitem[Faria et~al., 2010]{faria2010polar}
Faria, S.~H., Freitag, J., and Kipfstuhl, S. (2010).
\newblock Polar ice structure and the integrity of ice-core paleoclimate
  records.
\newblock {\em Quaternary Science Reviews}, 29(1-2):338--351.

\bibitem[Faria et~al., 2014]{faria2014microstructure}
Faria, S.~H., Weikusat, I., and Azuma, N. (2014).
\newblock The microstructure of polar ice. part i: Highlights from ice core
  research.
\newblock {\em Journal of Structural Geology}, 61:2--20.

\bibitem[Fischer et~al., 2021]{fischer2021ice}
Fischer, H., Blunier, T., and Mulvaney, R. (2021).
\newblock Ice cores: Archive of the climate system.
\newblock In {\em Glaciers and Ice Sheets in the Climate System}, pages
  279--325. Springer.

\bibitem[Fischer et~al., 2013]{fischer2013find}
Fischer, H., Severinghaus, J., Brook, E., Wolff, E., and Albert, M. (2013).
\newblock Where to find 1.5 million yr old ice for the ipics" oldest ice" ice
  core.
\newblock {\em Climate of the Past}, 9:2489--2505.

\bibitem[Florack et~al., 1996]{florack1996gaussian}
Florack, L., Romeny, B. T.~H., Viergever, M., and Koenderink, J. (1996).
\newblock The gaussian scale-space paradigm and the multiscale local jet.
\newblock {\em International Journal of Computer Vision}, 18(1):61--75.

\bibitem[Florack et~al., 1994]{florack1994linear}
Florack, L.~M., ter Haar~Romeny, B.~M., Koenderink, J.~J., and Viergever, M.~A.
  (1994).
\newblock Linear scale-space.
\newblock {\em Journal of Mathematical Imaging and Vision}, 4(4):325--351.

\bibitem[Gay and Weiss, 1999]{gay1999automatic}
Gay, M. and Weiss, J. (1999).
\newblock Automatic reconstruction of polycrystalline ice microstructure from
  image analysis: application to the epica ice core at dome concordia,
  antarctica.
\newblock {\em Journal of Glaciology}, 45(151):547--554.

\bibitem[Iizuka et~al., 2009]{iizuka2009constituent}
Iizuka, Y., Miyake, T., Hirabayashi, M., Suzuki, T., Matoba, S., Motoyama, H.,
  Fujii, Y., and Hondoh, T. (2009).
\newblock Constituent elements of insoluble and non-volatile particles during
  the last glacial maximum exhibited in the dome fuji (antarctica) ice core.
\newblock {\em Journal of Glaciology}, 55(191):552--562.

\bibitem[Kass et~al., 1988]{activecontours}
Kass, M., Witkin, A., and Terzopoulos, D. (1988).
\newblock Snakes: Active contour models.
\newblock {\em International Journal of Computer Vision}, 1(4):321--331.

\bibitem[Kerch et~al., 2015]{kerch2015laser}
Kerch, J., Spaulding, N., and Bohleber, P. (2015).
\newblock Laser ablation icp-ms on kcc microstructure-pilot study.

\bibitem[Kipfstuhl et~al., 2009]{kipfstuhl2009evidence}
Kipfstuhl, S., Faria, S.~H., Azuma, N., Freitag, J., Hamann, I., Kaufmann, P.,
  Miller, H., Weiler, K., and Wilhelms, F. (2009).
\newblock Evidence of dynamic recrystallization in polar firn.
\newblock {\em Journal of Geophysical Research: Solid Earth}, 114(B5).

\bibitem[Koenderink, 2021]{Koenderink2021}
Koenderink, J. (2021).
\newblock The structure of images: 1984–2021.
\newblock {\em Biological Cybernetics}.

\bibitem[Koenderink, 1984]{Koe84}
Koenderink, J.~J. (1984).
\newblock The structure of images.
\newblock {\em Biological Cybernetics}, 50:363--370.

\bibitem[Lindberg, 1994]{Lin94}
Lindberg, T. (1994).
\newblock {\em Scale-Space Theory in Computer Vision}.
\newblock Springer, Dordrecht.

\bibitem[Long et~al., 2015]{FCN15}
Long, J., Shelhamer, E., and Darrell, T. (2015).
\newblock Fully convolutional networks for semantic segmentation.
\newblock In {\em {IEEE} Conference on Computer Vision and Pattern Recognition,
  {CVPR} 2015, Boston, MA, USA, June 7-12, 2015}, pages 3431--3440. {IEEE}
  Computer Society.

\bibitem[Malladi et~al., 1995]{malladi1995shape}
Malladi, R., Sethian, J.~A., and Vemuri, B.~C. (1995).
\newblock Shape modeling with front propagation: A level set approach.
\newblock {\em IEEE transactions on pattern analysis and machine intelligence},
  17(2):158--175.

\bibitem[Morcillo et~al., 2020]{morcillo2020unravelling}
Morcillo, G., Faria, S.~H., and Kipfstuhl, S. (2020).
\newblock Unravelling antarctica’s past through the stratigraphy of a deep
  ice core: an image-analysis study of the epica-dml line-scan images.
\newblock {\em Quaternary International}, 566:6--15.

\bibitem[M{\"u}ller et~al., 2011]{muller2011direct}
M{\"u}ller, W., Shelley, J. M.~G., and Rasmussen, S.~O. (2011).
\newblock Direct chemical analysis of frozen ice cores by uv-laser ablation
  icpms.
\newblock {\em Journal of Analytical Atomic Spectrometry}, 26(12):2391--2395.

\bibitem[Ng et~al., 2002]{spectralclustering}
Ng, A., Jordan, M., and Weiss, Y. (2002).
\newblock On spectral clustering: {A}nalysis and an algorithm.
\newblock In Dietterich, T., Becker, S., and Ghahramani, Z., editors, {\em
  Advances in Neural Information Processing Systems}, volume~14, pages
  849--856. MIT Press.

\bibitem[Ng, 2020]{ng2020pervasive}
Ng, F.~S. (2020).
\newblock Pervasive diffusion of climate signals recorded in ice-vein ionic
  impurities.
\newblock {\em The Cryosphere Discussions}, pages 1--38.

\bibitem[Oyabu et~al., 2020]{oyabu2020compositions}
Oyabu, I., Iizuka, Y., Kawamura, K., Wolff, E., Severi, M., Ohgaito, R.,
  Abe-Ouchi, A., and Hansson, M. (2020).
\newblock Compositions of dust and sea salts in the dome c and dome fuji ice
  cores from last glacial maximum to early holocene based on ice-sublimation
  and single-particle measurements.
\newblock {\em Journal of Geophysical Research: Atmospheres},
  125(4):e2019JD032208.

\bibitem[Pavan and Pelillo, 2007]{pavan2007dominant}
Pavan, M. and Pelillo, M. (2007).
\newblock Dominant sets and pairwise clustering.
\newblock {\em IEEE Transactions on Pattern Analysis and Machine Intelligence},
  29(1):167--172.

\bibitem[Rempel et~al., 2001]{rempel2001possible}
Rempel, A., Waddington, E., Wettlaufer, J., and Worster, M. (2001).
\newblock Possible displacement of the climate signal in ancient ice by
  premelting and anomalous diffusion.
\newblock {\em Nature}, 411(6837):568--571.

\bibitem[Ren et~al., 2015]{FRCNN}
Ren, S., He, K., Girshick, R., and Sun, J. (2015).
\newblock Faster {R-CNN}: Towards real-time object detection with region
  proposal networks.
\newblock In Cortes, C., Lawrence, N., Lee, D., Sugiyama, M., and Garnett, R.,
  editors, {\em Advances in Neural Information Processing Systems}, volume~28.
  Curran Associates, Inc.

\bibitem[Romero et~al., 2020]{romero2020attentive}
Romero, D., Bekkers, E., Tomczak, J., and Hoogendoorn, M. (2020).
\newblock Attentive group equivariant convolutional networks.
\newblock In {\em International Conference on Machine Learning}, pages
  8188--8199. PMLR.

\bibitem[Ronneberger et~al., 2015]{UNET}
Ronneberger, O., Fischer, P., and Brox, T. (2015).
\newblock U-net: Convolutional networks for biomedical image segmentation.
\newblock In Navab, N., Hornegger, J., Wells, W.~M., and Frangi, A.~F.,
  editors, {\em Medical Image Computing and Computer-Assisted Intervention --
  MICCAI 2015}, pages 234--241, Cham. Springer International Publishing.

\bibitem[R{\"o}thlisberger et~al., 2000]{rothlisberger2000technique}
R{\"o}thlisberger, R., Bigler, M., Hutterli, M., Sommer, S., Stauffer, B.,
  Junghans, H.~G., and Wagenbach, D. (2000).
\newblock Technique for continuous high-resolution analysis of trace substances
  in firn and ice cores.
\newblock {\em Environmental Science \& Technology}, 34(2):338--342.

\bibitem[{\v{S}}ala et~al., 2020]{vsala2020analytical}
{\v{S}}ala, M., {\v{S}}elih, V.~S., Stremtan, C.~C., and van Elteren, J.~T.
  (2020).
\newblock Analytical performance of a high-repetition rate laser head (500 hz)
  for hr la-icp-qms imaging.
\newblock {\em Journal of Analytical Atomic Spectrometry}.

\bibitem[Shi and {Malik}, 2000]{NormalizedCut}
Shi, J. and {Malik}, J. (2000).
\newblock Normalized cuts and image segmentation.
\newblock {\em IEEE Transactions on Pattern Analysis and Machine Intelligence},
  22(8):888--905.

\bibitem[Sneed et~al., 2015]{sneed2015new}
Sneed, S.~B., Mayewski, P.~A., Sayre, W., Handley, M.~J., Kurbatov, A.~V.,
  Taylor, K.~C., Bohleber, P., Wagenbach, D., Erhardt, T., and Spaulding, N.~E.
  (2015).
\newblock New la-icp-ms cryocell and calibration technique for sub-millimeter
  analysis of ice cores.
\newblock {\em Journal of glaciology}, 61(226):233--242.

\bibitem[Spaulding et~al., 2017]{spaulding2017new}
Spaulding, N.~E., Sneed, S.~B., Handley, M.~J., Bohleber, P., Kurbatov, A.~V.,
  Pearce, N.~J., Erhardt, T., and Mayewski, P.~A. (2017).
\newblock A new multielement method for la-icp-ms data acquisition from glacier
  ice cores.
\newblock {\em Environmental science \& technology}, 51(22):13282--13287.

\bibitem[Svensson et~al., 2005]{svensson2005visual}
Svensson, A., Nielsen, S.~W., Kipfstuhl, S., Johnsen, S.~J., Steffensen, J.~P.,
  Bigler, M., Ruth, U., and R{\"o}thlisberger, R. (2005).
\newblock Visual stratigraphy of the north greenland ice core project
  (northgrip) ice core during the last glacial period.
\newblock {\em Journal of Geophysical Research: Atmospheres}, 110(D2).

\bibitem[ter Haar~Romeny, 2003]{ter2003front}
ter Haar~Romeny, B.~M. (2003).
\newblock {\em Front-end vision and multi-scale image analysis}.
\newblock Kluwer Academic Publ. Dordrecht.

\bibitem[Traversi et~al., 2009]{traversi2009sulfate}
Traversi, R., Becagli, S., Castellano, E., Marino, F., Rugi, F., Severi, M.,
  Angelis, M.~d., Fischer, H., Hansson, M., Stauffer, B., et~al. (2009).
\newblock Sulfate spikes in the deep layers of epica-dome c ice core: Evidence
  of glaciological artifacts.
\newblock {\em Environmental science \& technology}, 43(23):8737--8743.

\bibitem[Ueltzh{\"o}ffer et~al., 2010]{ueltzhoffer2010distribution}
Ueltzh{\"o}ffer, K.~J., Bendel, V., Freitag, J., Kipfstuhl, S., Wagenbach, D.,
  Faria, S.~H., and Garbe, C.~S. (2010).
\newblock Distribution of air bubbles in the edml and edc (antarctica) ice
  cores, using a new method of automatic image analysis.
\newblock {\em Journal of Glaciology}, 56(196):339--348.

\bibitem[van Elteren et~al., 2019]{van2019insights}
van Elteren, J.~T., {\v{S}}elih, V.~S., and {\v{S}}ala, M. (2019).
\newblock Insights into the selection of 2d la-icp-ms (multi) elemental mapping
  conditions.
\newblock {\em Journal of Analytical Atomic Spectrometry}, 34(9):1919--1931.

\bibitem[Van~Malderen et~al., 2017]{van2017imaging}
Van~Malderen, S.~J., Laforce, B., Van~Acker, T., Vincze, L., and Vanhaecke, F.
  (2017).
\newblock Imaging the 3d trace metal and metalloid distribution in mature wheat
  and rye grains via laser ablation-icp-mass spectrometry and micro-x-ray
  fluorescence spectrometry.
\newblock {\em Journal of Analytical Atomic Spectrometry}, 32(2):289--298.

\bibitem[Wang et~al., 2013]{wang2013fast}
Wang, H.~A., Grolimund, D., Giesen, C., Borca, C.~N., Shaw-Stewart, J.~R.,
  Bodenmiller, B., and G{\"u}nther, D. (2013).
\newblock Fast chemical imaging at high spatial resolution by laser ablation
  inductively coupled plasma mass spectrometry.
\newblock {\em Analytical chemistry}, 85(21):10107--10116.

\bibitem[Witkin, 1983]{witkin1983scale}
Witkin, A. (1983).
\newblock Scale-space filtering.
\newblock In {\em Proceedings of the Eighth International Joint Conference on
  Artificial Intelligence (IJCAI'83)}, volume~2, pages 1019--1022.

\bibitem[Zhou et~al., 2016]{zhou2015cnnlocalization}
Zhou, B., Khosla, A., A., L., Oliva, A., and Torralba, A. (2016).
\newblock {Learning Deep Features for Discriminative Localization.}
\newblock {\em CVPR}.

\end{thebibliography}

\end{document}